%% file: neurips_2026.tex
\documentclass{article}

\usepackage[preprint]{neurips_2026}

\usepackage[utf8]{inputenc} 
\usepackage[T1]{fontenc}    
\usepackage{hyperref}       
\usepackage{url}            
\usepackage{booktabs}       
\usepackage{amsfonts}       
\usepackage{nicefrac}       
\usepackage{microtype}      
\usepackage{xcolor}         
\usepackage{amsmath}
\usepackage{graphicx}

\title{ScaLe-INR: Scale and Learn Implicit Neural Representations}

\author{Buwaneka Epakanda, Athulya Ratnayake, Pandula Thennakoon, Mario De Silva \& Avishka Ranasinghe\\ 
\texttt{\{e19101,e19328,e18359,e19463,e18280\}@eng.pdn.ac.lk} \\
\And
Roshan Godaliyadda \& Parakrama Ekanayake \\
\texttt{roshang@eng.pdn.ac.lk \& mpb.ekanayake@ee.pdn.ac.lk} \\
}

\begin{document}

\maketitle

\input{content/0_Abstract}

\input{content/1_introduction}
\input{content/2_RelatedWork}

\input{content/3_methodology}
\input{content/4_results}
\input{content/6_conclusion}



\bibliographystyle{plainnat}
\bibliography{paper_refs}








\appendix

\input{content/appendix}



\end{document}

%% file: content/0_Abstract.tex
\begin{abstract}
Implicit Neural Representations (INRs) parameterized by multilayer perceptrons excel at modeling continuous signals. However, a key challenge persists as INRs fundamentally suffer from spectral bias and information cross-talk. When a single network attempts to capture multi-scale phenomena, high-frequency weight updates destructively interfere with the underlying low-frequency structural approximation. We introduce Scale and Learn INR (ScaLe-INR), a novel multi-branch architecture that resolves these limitations by explicitly matching the signal's frequency spectrum with the optimal operating region of the INR. Drawing upon the Fourier inverse scaling theorem we demonstrate that applying directional coordinate scaling expands a network's representational bandwidth along specific spatial axes. To mathematically enforce functional disentanglement and minimize task-specific information leakage between branches, we propose a Directional Edge Guidance Loss, a spatially-conditioned sparsity prior derived from ground-truth gradients. By constraining the high-frequency branches to act as strict, localized edge-filters, ScaLe-INR eliminates spectral cross-talk, accelerates convergence, and achieves high-fidelity signal reconstruction on complex multi-scale topologies. We evaluate ScaLe-INR across diverse reconstruction and inverse tasks, demonstrating substantial performance gains over existing state-of-the-art (SOTA) methods. The proposed architecture improves upon the nearest baselines by +5.16 dB in image reconstruction and +0.65 dB in image denoising. Furthermore, it achieve an impressive figure of 50.02 dB on audio reconstruction and 0.999 IOU(Intersection Over Union) on 3D reconstruction which beats the all SOTA models.

\end{abstract}

%% file: content/1_introduction.tex
\section{Introduction}

Implicit Neural Representations (INRs) represent signals as continuous functions parameterised by neural networks, learning a mapping from input coordinates to signal values rather than storing signals explicitly on a discrete grid~\citep{Where_Do_We_Stand_with_Implicit_Neural_Representations,jayasundara2026implicitneuralrepresentationssignal}. For images, this corresponds to mapping spatial coordinates to RGB intensities, while for audio, temporal coordinates are mapped to amplitudes. This continuous formulation enables resolution-independent signal modelling and has made INRs effective for image reconstruction, compression, super-resolution, novel view synthesis, and scientific signal representation.

Despite this flexibility, INRs exhibit spectral bias, where low-frequency components are learned more easily and earlier than high-frequency components~\citep{On-The-Spectral-Bias}. In image representation, this often leads to accurate reconstruction of smooth colour variations and coarse structures, while sharp edges, textures, corners, and fine local details remain difficult to capture. Since these high-frequency components are essential for perceptual and structural fidelity, improving the ability of INRs to represent them remains a central challenge in continuous signal representation.

Existing approaches address this limitation by modifying the INR itself, for example through activations, positional encodings, architectural changes, or training strategies~\citep{Where_Do_We_Stand_with_Implicit_Neural_Representations}. These methods aim to expand the frequency range that the network can directly model. In this work, we take a complementary perspective: instead of only adapting the INR to the signal spectrum, we investigate whether the signal spectrum can be transformed into a range that is more accessible to the INR. Motivated by the Fourier inverse scaling property, where scaling a signal in the signal domain inversely scales its frequency spectrum, we propose to use coordinate scaling to align relevant spectral components with the effective operating range of the INR.

We first study this idea using one-dimensional signals, where scaling offers a controlled way to examine how spectral content can be shifted into a more learnable range. These experiments show that scaling is beneficial only when appropriately matched: moderate scaling improves the representation of useful frequencies, while excessive scaling can over-compress the spectrum and degrade signal structure. Extending this principle to images, we apply scaling along the $x$ direction, the $y$ direction, or both, leading to a multi-branch INR architecture in which each branch receives a differently scaled coordinate representation. The unscaled branch captures low-frequency image structure, while the scaled branches focus on directional and joint high-frequency variations, enabling richer reconstruction than a standalone INR.

Since different branches may be more reliable in different regions depending on local frequency content and structural complexity, we introduce a confidence-based fusion strategy. This allows the network to adaptively weight branch contributions at each spatial location rather than treating all branches equally. To further encourage meaningful specialization, we introduce an edge-guided regularization term that weakly guides high-frequency branches toward edges and fine details, while allowing smoother regions to be represented primarily by low-frequency components. This regularization is gradually introduced during training so that it supports, rather than dominates, the primary reconstruction objective.

The main contributions of this work can be summarized as follows.

\begin{enumerate}
    \item We present a Fourier inverse scaling perspective for improving overall spectral representation in INRs. We show that signal-domain scaling can transform selected high-frequency components into a range that is more accessible to a SIREN-based INR, while also highlighting the need for controlled scaling.
    \item We propose a multi-branch INR framework for frequency-specialized signal representation. The framework is motivated by the observation that different scaled versions of a signal can make different frequency components more learnable for the INR by matching the relevant signal spectral content to the optimal operating region of the INR. For higher dimensional signals, this leads to multiple branches, which capture and model such specialized frequency components in a complementary manner. We further generalize this formulation to $N$-dimensional signals.
    \item We introduce confidence-based fusion and safe edge-guided regularization to support meaningful branch specialization. The fusion mechanism allows the model to adaptively combine the outputs of different frequency-specialized branches at a pixel level, enabling for detailed spectral information capture, that allows the proposed algorithm to significantly outperform the state-of-the-art.
\end{enumerate}

%% file: content/2_RelatedWork.tex
\section{Related Work}

\subsection{INRs and Spectral Bias}
Spectral bias is widely recognized as a key limitation of implicit neural representations (INRs), where neural networks tend to learn smooth, low-frequency structures more easily than rapidly varying details. Early coordinate-based models employing ReLU activations ~\citep{ReLU-NEURIPS2019} demonstrated the feasibility of representing signals continuously, but their limited spectral expressivity reduced their ability to reconstruct fine textures and high-frequency components. This tendency was systematically studied by~\citet{On-The-Spectral-Bias}, who showed that standard neural networks inherently favor low-frequency functions during optimization. To improve the frequency representation capability of INRs, several works introduced alternative activation mechanisms with richer spectral properties. Among these, SIREN~\citep{SIREN} employed periodic activations of the form $\phi(x)=\sin(\omega_0 x)$ enabling improved modeling of oscillatory patterns and stable gradient propagation. Subsequent approaches explored localized ~\citep{WIRE}~\citep{GAUSS-Beyond_Periodicity} or adaptive activation ~\citep{INCODE} designs to further enhance spectral coverage and reduce reconstruction artifacts. Following these different approaches, novel INRs are developed ~\citep{COSMO-INR},~\citep{FINER} , ~\citep{Fourier-features-let-networks-learn-high-Freq} , ~\citep{HOSC} , ~\citep{F_INR}.These developments collectively suggest that the activation function plays a critical role in determining the spectral characteristics and representation capacity of INRs.

\subsection{Multi-Scale and Frequency-Partitioned Architectures}

To address the limitations of single-branch INRs on complex topographies, recent literature has explored multi-scale partitioning strategies. One dominant approach relies on spatial partitioning, where the signal domain is divided into discrete hierarchical grids or trees ~\citep{ACRON}, ~\citep{Instant_NGP}. While these methods achieve rapid convergence by storing local features in trainable hash tables or octrees, they fundamentally sacrifice the memory efficiency and infinite continuous differentiability of pure MLP-based INRs, often introducing boundary artifacts at grid intersections.

An alternative approach focuses on frequency partitioning within purely continuous networks.~\citet{BACON} introduces a band-limited coordinate network that analytically models the signal at distinct frequency scales, while ~\citep{MINER}MINER routes coordinates through a Laplacian pyramid structure to capture multi-scale residuals. Further, parallel MLPs have been used in INR applications such as ~\citep{Cropable_INR}

Furthermore, the effect of scaling on the input coordinates is used in ~\citep{SIREN} for audio signals. However, the theoretical explanation of the scaling is not explored. ~\citet{IEEE_Journal} has explored different kernel transformations on the input coordinates and suggested that the linear transformation is better than non-linear kernels. However, that does not explain how the signal spectrum is changed for the learning process of INRs.

%% file: content/3_methodology.tex
\section{Methods}
\label{method}




\subsection{Fourier Scaling Perspective for INR Bandwidth Matching}

INRs are only capable of modeling a limited range of frequency components, as shown by ~\citet{On-The-Spectral-Bias}. Therefore, when the effective bandwidth of the signal exceeds the representational bandwidth of the INR, the model is unable to represent high frequency components.

Our major hypothesis is that this limitation can be reduced by transforming the high frequency components of the signal into a frequency range that is more compatible with the INR. If such components are scaled into the learnable bandwidth of the network, then the INR should be able to approximate them more effectively. 

if we enforce a single MLP INR to learn the signal, this will lead to suboptimal representation of both ends of the frequency spectrum. This is because unlike CNNs that use localized filters, MLPs are global function approximators and every parameter in the network has some influence over the entire spatial domain. therefore, we deploy multiple parallel sub-MLPs that specialize on sub-frequency bands of the signal.


We use the duality property of the Fourier transform to control the spectral distribution of a signal through scaling in the signal domain. Let $x(t)$ be a one dimensional signal with Fourier transform $X(\omega)$, such that $x(t) \longleftrightarrow X(\omega)$ If we scale the signal in the time domain as $x(\frac{t}{a})$, where $a \neq0$, its Fourier transform becomes $|a|X(a \omega)$.


This relationship shows that scaling in the signal domain produces an inverse scaling in the frequency domain. When the signal is stretched in the time domain, its frequency spectrum is compressed, causing higher frequency components to move towards the lower frequency region. Conversely, when the signal is compressed in the time domain, its frequency spectrum expands towards higher frequencies.

This property provides a simple and effective mechanism for modifying the spectral content observed by an INR. In particular, we can use signal domain scaling to map high frequency components into a lower frequency range that lies within the effective learning bandwidth of the network. This directly supports our hypothesis, since the transformed signal becomes easier for the INR to approximate while still preserving information from the original high frequency content.

\subsection{One-Dimensional Analysis of Coordinate Scaling}

\begin{figure}
  \centering
  \includegraphics[width=0.8\textwidth]{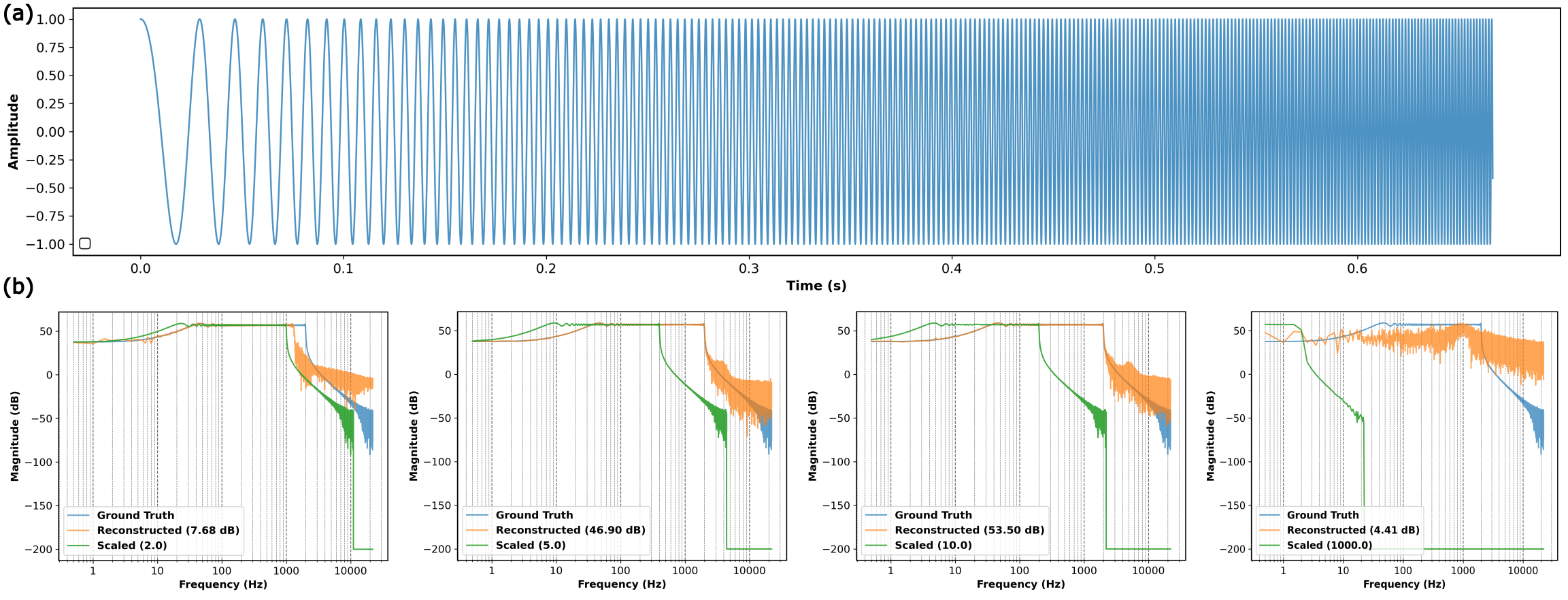}
  \caption{Effect of signal scaling on SIREN reconstruction of a chirp signal. Moderate scaling improves the match between the reconstructed and ground-truth spectra by shifting high-frequency content into a more learnable range, while excessive scaling degrades reconstruction by over-compressing the spectrum.}
  \label{fig:chirp}
\end{figure}

To examine frequency compression, we train a standard SIREN-based INR to reconstruct a one-dimensional chirp signal, whose time-varying frequency content makes it useful for evaluating reconstruction across different spectral regions. As shown in Fig.~\ref{fig:chirp}, we apply different scaling factors and compare the ground truth and reconstructed signals in both the time and frequency domains.

Fig.~\ref{fig:chirp} shows that increasing the scaling factor initially improves the reconstruction of high-frequency components, both in the time domain and in the corresponding frequency spectrum. However, this improvement saturates and eventually degrades when the scaling factor becomes too large, as excessive stretching over-compresses the spectrum and makes important spectral structure less distinguishable. These results support our hypothesis that scaling can make high-frequency content more accessible to the INR, while also demonstrating the need for controlled scaling before extending the approach to higher-dimensional signals such as images.

\subsection{Scale-Conditioned Multi-Branch INR Architecture}


\begin{figure}
  \centering
  \includegraphics[width=0.9\textwidth]{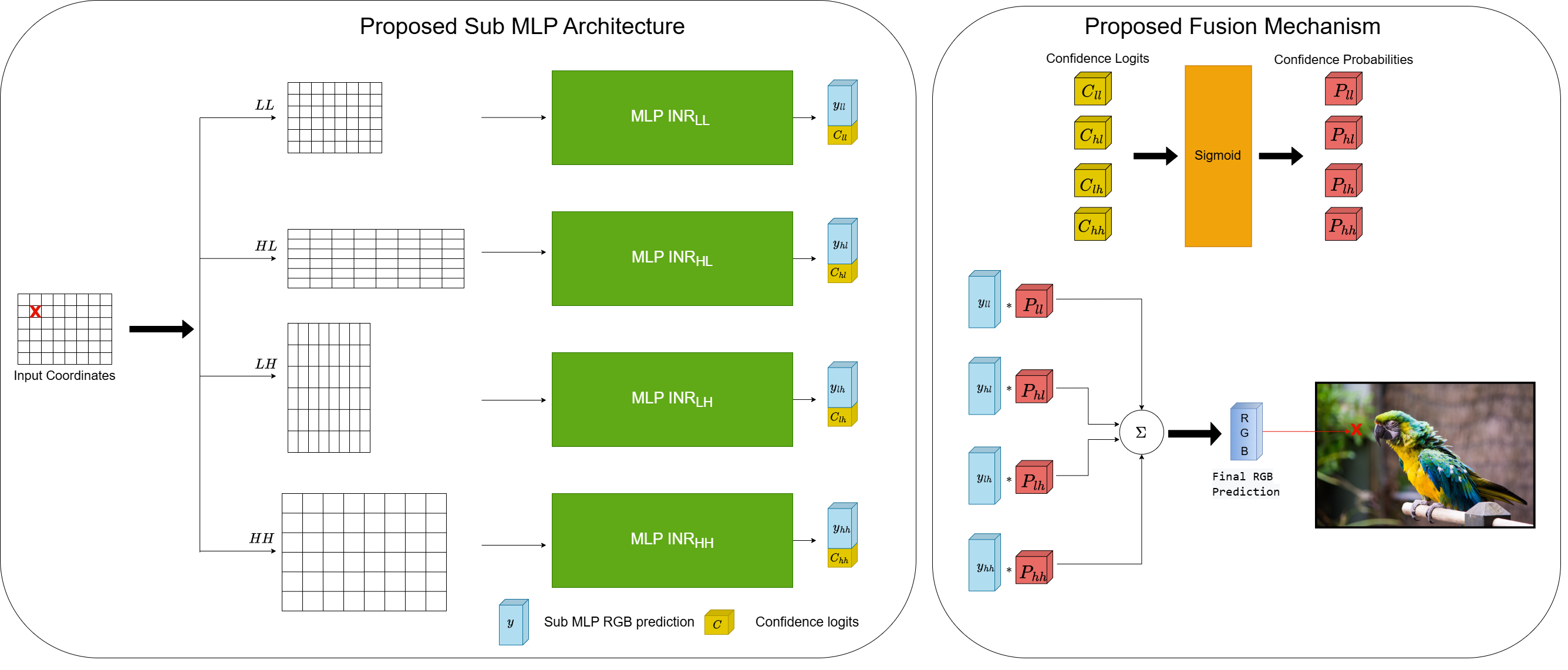}
  \caption{ScaLe-INR architecture for image-related tasks. There are 4 parallel branches ($LL$,$HL$,$LH$ and $HH$) that scale in different directions.}
  \label{fig:overall_architecture}
\end{figure}

Let an image be represented as a continuous function $\boldsymbol{I}(x,y) \in \mathbb{R}^3$, where $(x,y)$ denotes the spatial coordinate and $\boldsymbol{I}(x,y)$ gives the RGB value at that coordinate. Since an image has frequency content along both horizontal and vertical directions, we use a four-branch architecture to separately model different frequency regions. The four branches are denoted by $LL, HL, LH,\text{ and } HH$. Here $L$ denotes a low frequency component and $H$ denotes a high frequency component.The branches can be interpreted as follows:
\begin{itemize}
    \item $LL$: low frequency in both $x$ and $y$
    \item $HL$:high frequency in $x$ and low frequency in $y$
    \item $LH$:low frequency in $x$ and high frequency in $y$
    \item $HH$:high frequency in both $x$ and $y$
\end{itemize}

The justification for the existence of the branch that attempts to capture simultaneous high frequency variations along both $x$ and $y$ directions, in spite there being separate branches over the $x$ and $y$ directions separately is to leverage the non-linear behavior of the INR to optimally capture edge information that is resultant of simultaneous variation along $x$ and $y$ directions.

\begin{equation}
    \label{eq:superposition}
    f_{\theta}(x\odot y) \neq f_\theta(x) \odot f_\theta(y)
\end{equation}

The inherent non-linearity of the INR as would be discussed in Section~\ref{results} enables more intricate edge information to be captured by the $HH$ branch that are not obtainable through a linear operation of the $LH$ and $HL$ branches.

The $LL$ branch is expected to capture slowly varying image information, such as global colour structure, smooth intensity variations, and coarse textures. The $HL$ and $LH$ branches capture directional high frequency information along the horizontal and vertical spatial directions, respectively. The $HH$ branch captures components that vary rapidly along both directions, such as corners, fine structures, and highly localized details.

For each spatial coordinate $(x,y)$ each branch produces two outputs: an RGB prediction and a confidence logit. Let the RGB prediction of branch $b$ be $\boldsymbol{I}(x,y) \in \mathbb{R}^3$, and let its confidence logit be $c_b(x,y)$, where $b \in \{LL, HL, LH, HH\}$.

The confidence logits from all branches are concatenated and passed through a softmax operation to obtain normalized confidence weights:

\begin{equation}
    P_b(x,y) = \frac{\exp(c_b(x,y))}{\sum_{j\in \{LL, HL, LH, HH\}} \exp(c_j(x,y))}
\end{equation}



This allows the model to assign a spatially varying contribution to each branch. In regions where low frequency information is dominant, the LL branch can receive a larger weight. In regions containing edges, fine textures, or directional variations, the high frequency branches can contribute more strongly.

We define the contribution of each branch as the product of its RGB prediction and its confidence weight:, and the final reconstructed RGB value is then obtained by adding the confidence weighted contributions from all branches:

\begin{equation}
    \hat{\boldsymbol{I}}(x,y) = \sum_{b \in \{LL, HL, LH, HH\}} P_b(x,y)\hat{\boldsymbol{I}}_b(x,y)
\end{equation}

We refer to this mechanism as confidence based fusion. Instead of treating all branch outputs equally, the model learns how much each branch should contribute at each pixel location. As a result, the architecture combines frequency specialized representations while allowing the final prediction to be adaptively formed according to the local structure of the image. The overall architecture is given in Fig.~\ref{fig:overall_architecture}

\subsection{Training Objective and Edge-Guided Regularisation}

We train the proposed model using a reconstruction objective together with a safe edge-guidance term. The total loss used in the main mini-batch loop is defined as

\begin{equation}
    \mathcal{L}_{\mathrm{Total}} = \mathcal{L}_{\mathrm{recon}} + \mathcal{L}_{\mathrm{edge,safe}}
\end{equation}

where $\mathcal{L}_{\mathrm{edge,safe}} = \mathrm{safe}\left(\lambda_{\mathrm{edge}}(t) \mathcal{L}_{\mathrm{edge}}\right)$.

Here, $\mathcal{L}_{\mathrm{recon}}$ denotes the reconstruction loss, and $\mathcal{L}_{\mathrm{edge}}$ denotes the directional edge-guidance loss. The scalar $\lambda_{\mathrm{edge}}(t)$ is a scheduled weighting factor that controls the strength of the edge-guidance term at training step $t$. The function $\mathrm{safe}(\cdot)$ limits the weighted edge term so that it remains an auxiliary regularizer and does not dominate the reconstruction objective.

The reconstruction loss is defined as

\begin{equation}
\mathcal{L}_{\mathrm{recon}}
=
\mathbb{E}_{(x,y)\sim \mathcal{U}(\Omega_d)}
\left[
\left\|
\boldsymbol{\hat{I}}(x,y)-\boldsymbol{I}(x,y)
\right\|_2^2
\right]
\end{equation}

where $\boldsymbol{I}(x,y)\in\mathbb{R}^3$ is the ground-truth RGB value and $\boldsymbol{\hat{I}}(x,y)\in\mathbb{R}^3$ is the predicted RGB value at coordinate $(x,y)$. The expectation is taken over coordinates sampled uniformly from the discrete image domain $\Omega_d$, and in practice is estimated using the sampled mini-batch of pixel coordinates. Since reconstruction quality is evaluated using PSNR, which is a monotonic function of the mean squared error, minimizing $\mathcal{L}_{\mathrm{recon}}$ is directly aligned with the main optimization goal.

The edge-guidance loss is introduced to encourage meaningful branch specialization. The high-frequency branches should contribute more strongly near image structures such as edges and fine details, while their influence should be reduced in smooth regions. Therefore, we penalize the high-frequency branch contributions in regions where the corresponding directional edge masks are weak:

\begin{equation}
    \mathcal{L}_{\mathrm{edge}} = \mathbb{E} \left[(1-M_x)|C_{HL}|\right] +\mathbb{E} \left[(1-M_y)|C_{LH}|\right] + \mathbb{E} \left[(1-M_{all})|C_{HH}|\right] + \beta_{\mathrm{conf}}\mathcal{L}_{\mathrm{conf}}
\end{equation}

Here, $M_x$, $M_y$, and $M_{all}$ are the directional edge masks computed from Sobel gradients of the ground truth. The terms $C_{HL}$, $C_{LH}$, and $C_{HH}$ denote the confidence-weighted contributions of the corresponding high-frequency branches:

\begin{equation}
    C_b(x,y) = P_b(x,y)\boldsymbol{\hat{I}}_b(x,y), \qquad b \in \{HL,LH,HH\}
\end{equation}

This means that the edge loss is applied to the actual contribution of each branch to the final reconstruction, rather than to the raw branch output alone. This is consistent with the confidence fusion mechanism, since a branch affects the final prediction only through its weighted contribution.

The confidence regularization term is given by

\begin{equation}
    \mathcal{L}_{\mathrm{conf}} = \mathbb{E}\left[(1-M_x)P_{HL}\right] + \mathbb{E}\left[(1-M_y)P_{LH}\right] + \mathbb{E}\left[(1-M_{all})P_{HH}\right],
\end{equation}

where $\beta_{\mathrm{conf}}$ controls the strength of this regularization. This term gently discourages the model from assigning high confidence to high-frequency branches in regions where strong edge structures are not present.

$\mathcal{L}_{\mathrm{edge}}$ provides a mechanism for encouraging pixel-level branch specialization by aligning each branch’s confidence with the local spectral structure of the signal. In particular, it promotes higher confidence for branches whose frequency specialization matches the dominant spatial variation at a given location, while discouraging branches from contributing in regions outside their intended domain of specialization. As a result, the confidence-based fusion mechanism can adaptively assign greater weight to the most relevant branch at each pixel, leading to a more structured and frequency-aware reconstruction.

The scheduled weighting factor $\lambda_{\mathrm{edge}}(t)$ allows the model to first learn a stable image reconstruction before the edge-guidance term becomes active. After the scheduled warm-up period, the edge term is gradually introduced. To ensure that this auxiliary term does not overpower the reconstruction loss, the safe scaling operation constrains its effective contribution as $\lambda_{\mathrm{edge}}(t)\mathcal{L_{\mathrm{edge}}} \leq \alpha\mathcal{L}_{\mathrm{recon}}$, where $\alpha$ is the edge cap ratio. Thus, the total objective preserves the reconstruction loss as the dominant term, while using the edge-guidance loss to organize the branch contributions in a structurally meaningful way.

The overall architecture generalized for an $N$-dimensional signal is given in the Appendix~\ref{app:nd_generalisation}

%% file: content/4_results.tex
\section{Experiments \& Results}
\label{results}

We implement all experiments in PyTorch and evaluate on an NVIDIA RTX 6000 Ada GPU with 48 GB memory using the Adam optimizer. All tasks share a unified pipeline except 3D occupancy and audio reconstruction due to different input output formats, where occupancy uses 3D coordinates to a single output and audio follows a one to one mapping.

Each model uses 3-layer MLPs with hidden dimension 256 and four parallel branches, except occupancy reconstruction which uses eight branches. The high frequency multiplier is set to 4.0 based on ablation studies and reduced for low frequency tasks. Fusion temperature starts at 1.0 to encourage exploration in confidence based softmax weighting and is reduced during training for more deterministic selection. The edge cap ratio is fixed at 0.05 to limit edge loss contribution to at most 5 percent of reconstruction loss.

\subsection{Image Representation}
\label{res:image}

\textbf{Data.} We conducted our experiments on the Kodak~\citep{Kodak} dataset, comprising of 24 lossless images. The images were trained at their native resolution. Reconstruction performance was evaluated using PSNR as a metric to quantify the error against the ground truth image.

\begin{figure*}[h]
    \centering
    \includegraphics[width=0.9\linewidth]{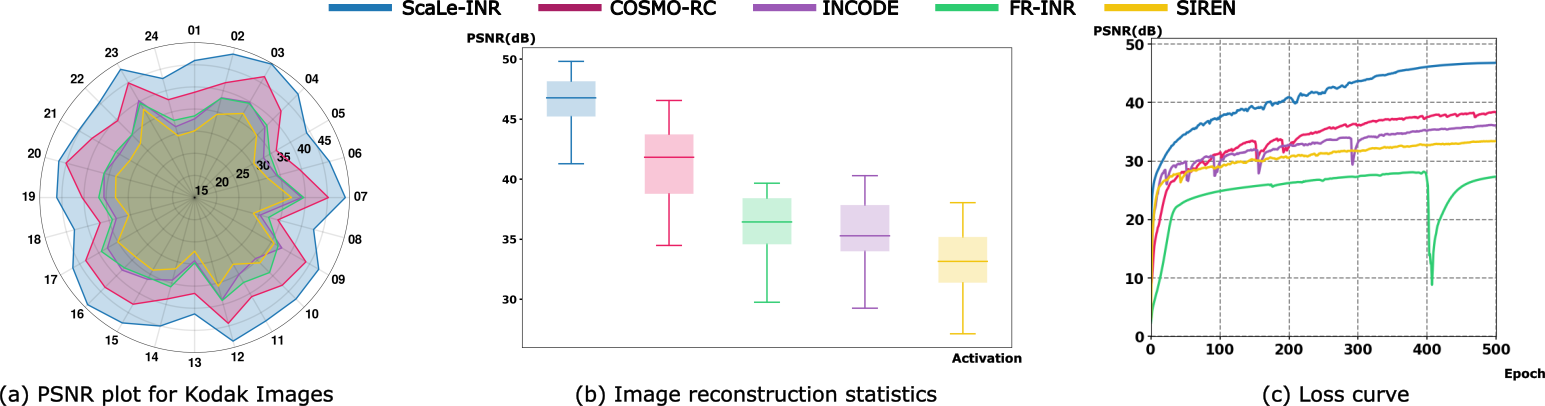}
    \caption{ScaLe-INR performance analysis on Kodak image reconstruction}
    \label{fig:kodak__loss_plot}
\end{figure*}

\textbf{Results.} The performance of ScaLe-INR in comparison with state-of-the-art (SOTA) methods on the Kodak dataset is presented in Fig.~\ref{fig:kodak__loss_plot}(a) The results demonstrate that ScaLe-INR consistently achieves superior reconstruction performance, attaining an average PSNR of 46.4 dB and outperforming the nearest competing method, COSMO-INR~\citep{COSMO-INR}, by 5.16 dB.


The performance statistics for each model are presented in Fig.~\ref{fig:kodak__loss_plot}(b). Furthermore, the loss curves shown in Fig.~\ref{fig:kodak__loss_plot}(c) demonstrate that ScaLe-INR converges significantly faster than the other models while maintaining stable optimization behavior throughout training.


\subsection{Image Denoising}
\label{res:denoise}

\textbf{Data.}
To evaluate robustness against noisy input signals, we utilize the \textit{Parrot} image from the DIV2K dataset~\citep{DIV2K}. Photon noise is simulated by applying independent Poisson random variables to the ground truth pixel intensities, followed by an additional additive noise component to increase corruption complexity.

\textbf{Implementation Details.}
The original image is downscaled by a factor of $1/2$ prior to training. Noise levels are configured using a photon noise factor of $4 \times 10^{1}$ and an additive SNR noise component of $2$, resulting in a degraded input with PSNR of $16.73$~dB. Models are trained for 500 epochs using a batch size of $256 \times 256$, a learning rate of $1.5 \times 10^{-4}$, and a decay factor of $0.1$. The training requires approximately 3700~MB of GPU memory. In accordance with the duality property in Section~\ref{method}, a frequency scaling factor of $0.5$ is used.

\begin{figure}[h]
    \centering
    \includegraphics[width=0.9\textwidth]{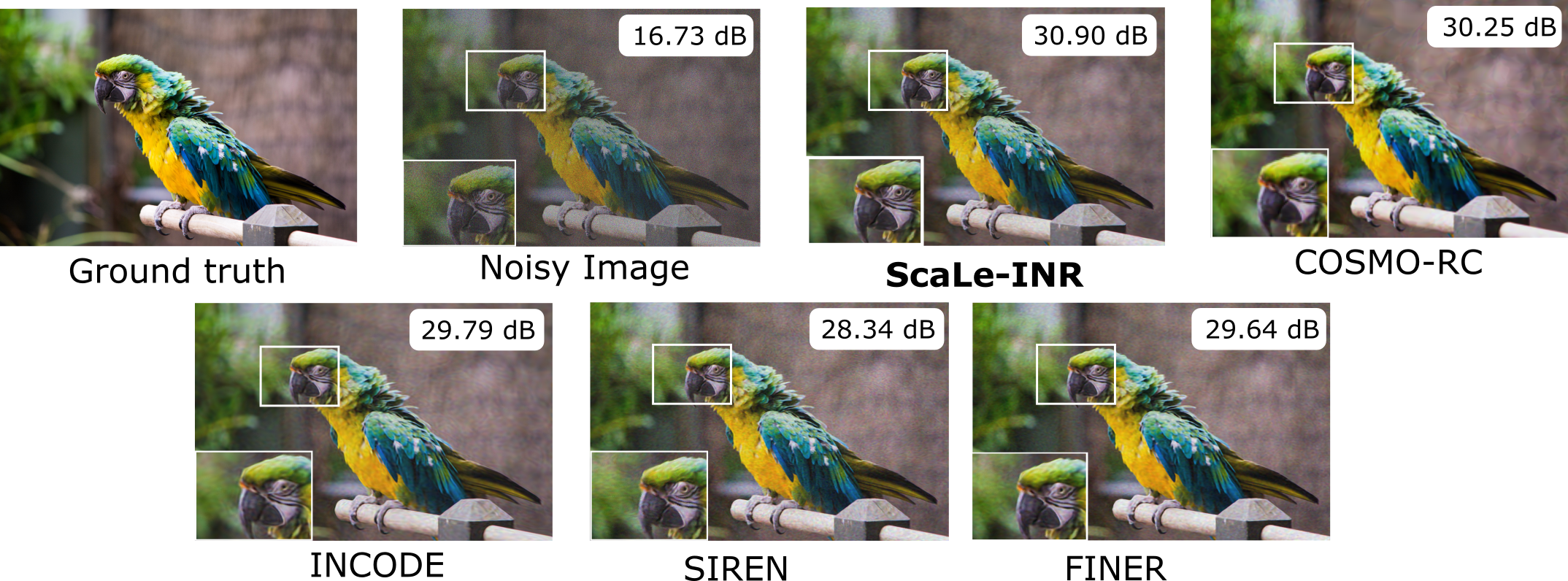}
    \caption{Qualitative denoising results comparing ScaLe-INR against existing approaches.}
    \label{fig:denoise}
\end{figure}

\textbf{Results.}
As shown in Fig.~\ref{fig:denoise}, ScaLe-INR achieves a PSNR of $30.90$~dB, outperforming COSMO-RC while maintaining superior structural fidelity. Notably, ScaLe-INR preserves fine anatomical details in challenging regions such as the beak, where competing methods tend to oversmooth or introduce artifacts. This indicates improved robustness to high-frequency noise and better edge preservation under severe corruption.

\subsection{Image Super-Resolution}
\label{res:supeRes}

\textbf{Data.}
We use a high-resolution image from the DIV2K dataset~\citep{DIV2K} with resolution $1356 \times 2040 \times 3$. Low-resolution inputs are generated using downsampling factors of $1/2$, $1/4$, and $1/6$, corresponding to $2\times$, $4\times$, and $6\times$ super-resolution tasks.

\textbf{Implementation Details.}
Models are trained for 500 epochs using a batch size of $256 \times 256$, learning rate $9 \times 10^{-4}$, and decay factor $0.1$. The architecture requires approximately 2250~MB GPU memory. Following Section~\ref{method}, frequency scaling factors of $0.3$, $0.5$, and $0.9$ are used for $2\times$, $4\times$, and $6\times$ settings respectively.

\textbf{Results.}
Table~\ref{Table:SuperRes} shows that ScaLe-INR consistently achieves the best performance across all scales. The largest gain is observed at $6\times$, where improved SSIM ($0.85$) indicates better structural preservation under severe downsampling. Furthermore, ScaLe-INR improves over COSMO-RC at all scales, with a notable gain of $0.40$ dB at $2\times$, demonstrating the effectiveness of frequency-aware modeling.

\begin{table}[h]
    \centering
    \begin{tabular}{lcccccc}
        \toprule
        Methods & \multicolumn{2}{c}{$2\times$} & \multicolumn{2}{c}{$4\times$} & \multicolumn{2}{c}{$6\times$} \\
        \cmidrule(lr){2-3}
        \cmidrule(lr){4-5}
        \cmidrule(lr){6-7}
        & PSNR & SSIM & PSNR & SSIM & PSNR & SSIM \\
        \midrule
        ReLU+PE   & 32.80 & 0.91 & 28.89 & 0.87 & 26.29 & 0.83 \\
        SIREN     & 32.26 & 0.90 & 29.62 & 0.87 & 27.31 & 0.81 \\
        INCODE    & 32.83 & 0.90 & 29.96 & 0.85 & 26.63 & 0.78 \\
        FINER     & 32.94 & 0.91 & 29.75 & 0.84 & 27.02 & 0.80 \\
        COSMO-RC  & 34.03 & 0.96 & 30.42 & 0.95 & 27.66 & 0.83 \\
        \midrule
        ScaLe-INR & 34.43 & 0.97 & 30.48 & 0.95 & 27.68 & 0.85 \\
        \bottomrule
    \end{tabular}
    \vspace{2mm}
    \caption{Comparison of super-resolution performance against existing SOTA methods.}
    \label{Table:SuperRes}
\end{table}






\subsection{3D Occupancy}
\label{res:occupancy}

\textbf{Data.}
We use the \textbf{Lucy} dataset and convert meshes into a $512^3$ occupancy grid with binary labels indicating occupied and empty voxels.

\textbf{Implementation Details.}
Training is conducted for 200 epochs using $10^6$ coordinate samples per batch, learning rate $1 \times 10^{-4}$, and decay factor $0.2$. Due to volumetric complexity, training requires approximately 20~GB GPU memory. Eight frequency pipelines are used to model full $\{H,L\}$ combinations across three spatial axes.

\textbf{Results.}
ScaLe-INR achieves the highest IoU ($0.999$), outperforming COSMO-RC ($0.995$), FINER ($0.994$), INCODE ($0.993$), and SIREN ($0.992$). The results demonstrate improved geometric fidelity and sharper boundary reconstruction, confirming the effectiveness of frequency scaling for 3D structure modeling.

\begin{figure}[h]
    \centering
    \includegraphics[width=0.8\textwidth]{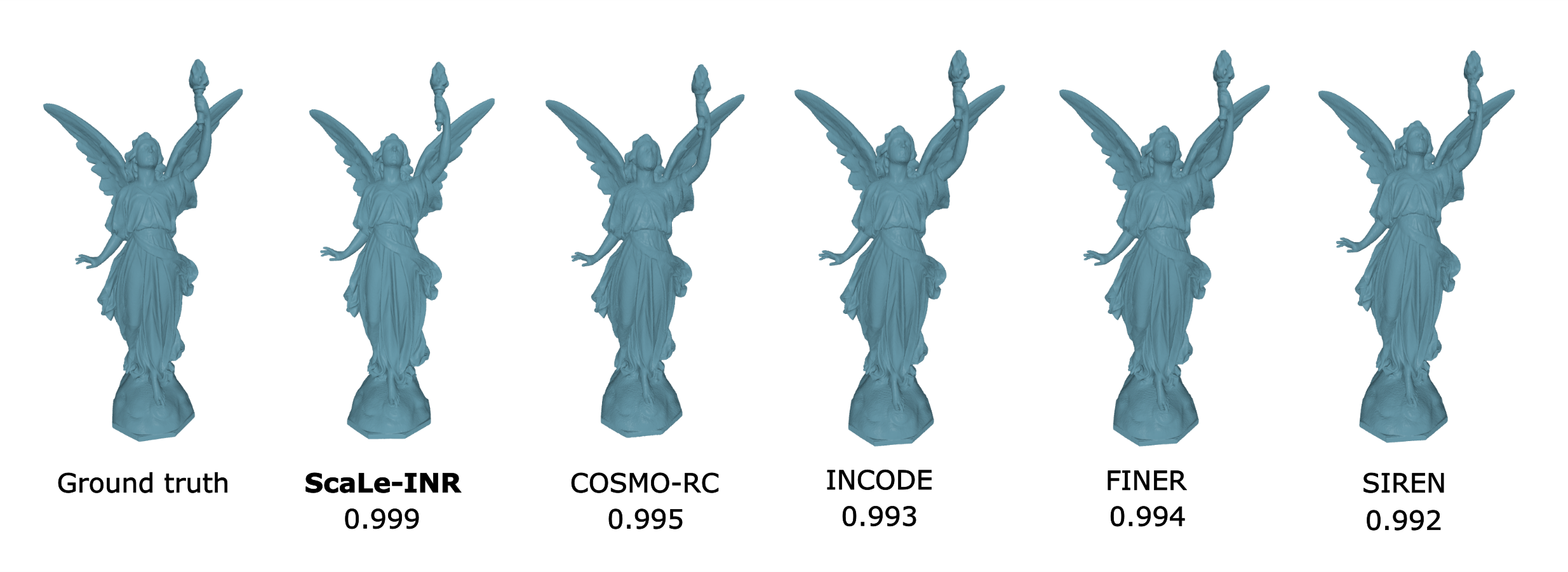}
    \caption{3D occupancy reconstruction results obtained using ScaLe-INR.}
    \label{fig:occupancy}
\end{figure}

\subsection{Audio Reconstruction}
\label{res:audio}
\textbf{Data.}
We use the first 7 seconds of Bach's Cello Suite No.~1 to evaluate audio reconstruction performance in terms of PSNR.

\textbf{Implementation Details.}
ScaLe-INR employs two parallel MLP branches modeling low and high frequency components. Frequency scaling is set to $100$ for low-frequency and $400$ for high-frequency components following prior work. The model is trained for 1000 epochs for fair comparison with INCODE.

\textbf{Results.}
ScaLe-INR achieves the best performance with a PSNR of $50.02$~dB, outperforming INCODE ($49.10$~dB), SIREN ($37.92$~dB), Gaussian features ($38.50$~dB), and ReLU+PE ($22.99$~dB). This demonstrates strong effectiveness in modeling high-fidelity temporal signals.


%% file: content/6_conclusion.tex
\section{Conclusion}

We presented Scale and Learn Implicit Neural Representations (ScaLe-INR), a novel multi-branch architecture designed to overcome the spectral bias and information cross-talk inherent in standard INRs. By conceptualizing coordinate networks as a multi-resolution continuous filter bank, we demonstrated that directional coordinate scaling inversely scale the representational bandwidth of the network along specific spatial axes. To enforce strict functional disentanglement, we introduced the Directional Edge Guidance Loss, which imposes a spatially-conditioned sparsity prior derived from ground-truth gradients. This structural prior ensures that the high-frequency branches act exclusively as localized directional edge filters, preventing them from corrupting the global low-frequency manifold. Our results confirm that by explicitly partitioning the signal's frequency spectrum and guiding the network with edge guidance loss, ScaLe-INR effectively eliminates information cross-talk, significantly accelerates convergence, and achieves high-fidelity signal reconstruction on complex multi-scale topologies. We believe our work marks a major milestone in this research domain and future researchers will benefit from our findings. In future, we aim to extend this directional scaling paradigm to higher-dimensional tasks, including dynamic scene reconstruction and neural radiance fields.

%% file: content/appendix.tex

\newpage

\section{Generalisation of the Proposed Method to $N$-Dimensional Signals}
\label{app:nd_generalisation}

This appendix shows that the multiresolution implicit neural representation proposed in Section~\ref{method} can be extended naturally from a two-dimensional image domain to a general $N$-dimensional signal domain. The construction preserves the core components of the proposed method: axis-wise frequency-scaled branches, confidence-weighted fusion, reconstruction supervision, directional edge guidance, and local gradient consistency.

\subsection{Two-Dimensional Formulation}

In the main implementation, the input is a two-dimensional RGB signal
\begin{equation}
    s:\Omega \subset \mathbb{R}^{2} \longrightarrow \mathbb{R}^{3},
\end{equation}
where $\mathbf{x}=(x_1,x_2)\in\Omega$ denotes a spatial coordinate and
$s(\mathbf{x})\in\mathbb{R}^{3}$ denotes the RGB value at that coordinate.

The model learns an approximation
\begin{equation}
    \hat{s}_{\theta}(\mathbf{x}) \approx s(\mathbf{x}).
\end{equation}

In the two-dimensional case, the branch set is $\{LL,HL,LH,HH\}$, where each symbol indicates whether the corresponding coordinate axis is treated as low-frequency or high-frequency.

\subsection{General $N$-Dimensional Signal Domain}

Let
\begin{equation}
    s:\Omega \subset \mathbb{R}^{N} \longrightarrow \mathbb{R}^{c}
\end{equation}
be an $N$-dimensional signal with $c$ channels. Here,
\begin{equation}
    \mathbf{x}=(x_1,x_2,\dots,x_N)\in\Omega
\end{equation}
denotes an $N$-dimensional coordinate, and $s(\mathbf{x})\in\mathbb{R}^{c}$ denotes the corresponding signal value.

The implicit neural representation is written as
\begin{equation}
    \hat{s}_{\theta}:\Omega \subset \mathbb{R}^{N} \longrightarrow \mathbb{R}^{c}.
\end{equation}

Although the neural network can be evaluated on $\mathbb{R}^{N}$, it is trained and evaluated on the signal domain $\Omega$ or on a finite sampled subset of it.

\subsection{Generalised Branch Structure}

For an $N$-dimensional signal, define the branch set as
\begin{equation}
    \mathcal{B}_{N}=\{L,H\}^{N}.
\end{equation}
Thus, each branch
\begin{equation}
    b=(b_1,b_2,\dots,b_N)\in\mathcal{B}_{N}
\end{equation}
is an $N$-tuple with
\begin{equation}
    b_i\in\{L,H\}, \qquad i=1,2,\dots,N.
\end{equation}

The two-dimensional case is recovered as
\begin{equation}
    \mathcal{B}_{2}
    =
    \{L,H\}^{2}
    =
    \{LL,HL,LH,HH\}.
\end{equation}

Since each of the $N$ coordinate axes has two possible frequency states, the number of branches is
\begin{equation}
    |\mathcal{B}_{N}| = 2^{N}.
\end{equation}

This establishes that the proposed branch construction extends directly from four branches in two dimensions to $2^N$ branches in $N$ dimensions.

\subsection{Axis-Wise Frequency Scaling}

For each branch $b\in\mathcal{B}_{N}$, define an axis-wise frequency-scaling vector
\begin{equation}
    \boldsymbol{\alpha}_{b}
    =
    (\alpha_{b_1},\alpha_{b_2},\dots,\alpha_{b_N}),
\end{equation}
where
\begin{equation}
    \alpha_{b_i}\in\{\alpha_L,\alpha_H\},
    \qquad
    0<\alpha_L<\alpha_H.
\end{equation}

Here, $\alpha_L$ denotes the low-frequency coordinate scaling factor and $\alpha_H$ denotes the high-frequency coordinate scaling factor. For example, in a three-dimensional signal,
\begin{equation}
    \boldsymbol{\alpha}_{LHL}
    =
    (\alpha_L,\alpha_H,\alpha_L).
\end{equation}

The input received by branch $b$ is therefore
\begin{equation}
    \boldsymbol{\alpha}_{b}\odot\mathbf{x},
\end{equation}
where $\odot$ denotes element-wise multiplication. This axis-wise scaling allows different branches to specialise in different frequency patterns across the coordinate axes.

\subsection{Branch Output and Confidence Logit}

Each branch $b\in\mathcal{B}_{N}$ is represented by a coordinate-based neural function
\begin{equation}
    f_b:\mathbb{R}^{N}\longrightarrow\mathbb{R}^{c+1}.
\end{equation}

For a coordinate $\mathbf{x}$, the branch output is written as
\begin{equation}
    f_b(\boldsymbol{\alpha}_{b}\odot\mathbf{x})
    =
    \left[
    \mathbf{y}_{b}(\mathbf{x}),
    c_b(\mathbf{x})
    \right],
\end{equation}
where
\begin{equation}
    \mathbf{y}_{b}(\mathbf{x})\in\mathbb{R}^{c}
\end{equation}
is the branch-specific signal prediction, and
\begin{equation}
    c_b(\mathbf{x})\in\mathbb{R}
\end{equation}
is the corresponding confidence logit.

Thus, each branch produces both a candidate reconstruction and a scalar confidence logit for that coordinate.

\subsection{Confidence-Weighted Fusion}

The confidence logits are normalised across all branches using a softmax function. For branch $b\in\mathcal{B}_{N}$, the confidence weight is defined as
\begin{equation}
    P_b(\mathbf{x})
    =
    \frac{
    \exp\left(c_b(\mathbf{x})/\tau\right)
    }{
    \sum_{b'\in\mathcal{B}_{N}}
    \exp\left(c_{b'}(\mathbf{x})/\tau\right)
    },
\end{equation}
where $\tau>0$ is the softmax temperature.

The weights satisfy
\begin{equation}
    P_b(\mathbf{x})\geq 0,
    \qquad
    \sum_{b\in\mathcal{B}_{N}}P_b(\mathbf{x})=1.
\end{equation}

The final reconstruction is then given by the convex combination
\begin{equation}
    \hat{s}_{\theta}(\mathbf{x})
    =
    \sum_{b\in\mathcal{B}_{N}}
    P_b(\mathbf{x})\mathbf{y}_{b}(\mathbf{x}).
\end{equation}

Therefore, the model adaptively combines the outputs of all frequency-scaled branches at each coordinate.

\subsection{Generalised Loss Functions}

\subsubsection{Reconstruction Loss}

For a discretely sampled signal defined on a finite coordinate set
\begin{equation}
    \Omega_d=\{\mathbf{x}_j\}_{j=1}^{|\Omega_d|}\subset\Omega,
\end{equation}
the reconstruction loss is defined as
\begin{equation}
    \mathcal{L}_{\mathrm{recon}}
    =
    \frac{1}{|\Omega_d|}
    \sum_{\mathbf{x}\in\Omega_d}
    \left\|
    \hat{s}_{\theta}(\mathbf{x})-s(\mathbf{x})
    \right\|_2^2.
\end{equation}

This is the natural $N$-dimensional extension of the pixel-wise mean squared error used in the two-dimensional implementation.

For a continuous signal domain, the analogous objective may be written as
\begin{equation}
    \mathcal{L}_{\mathrm{recon}}
    =
    \frac{1}{|\Omega|}
    \int_{\Omega}
    \left\|
    \hat{s}_{\theta}(\mathbf{x})-s(\mathbf{x})
    \right\|_2^2
    \,d\mathbf{x},
\end{equation}
provided that the integral is well defined.

\subsubsection{Directional Edge Masks}

To extend the directional edge-guidance mechanism to $N$ dimensions, we define one directional edge mask per coordinate axis. For a differentiable continuous signal, the edge strength along the $i$-th coordinate axis may be defined as
\begin{equation}
    M_i(\mathbf{x})
    \propto
    \left\|
    \frac{\partial s(\mathbf{x})}{\partial x_i}
    \right\|_2,
    \qquad
    i=1,2,\dots,N.
\end{equation}

Since $s(\mathbf{x})\in\mathbb{R}^{c}$, the partial derivative
$\partial s(\mathbf{x})/\partial x_i$ is channel-valued, and its Euclidean norm gives a scalar directional edge magnitude.

For discrete signals, these masks may be approximated using finite-difference operators, derivative filters, or Sobel-type filters when such filters are appropriate for the signal dimension. After computing the directional magnitudes, a normalisation and optional softening operation may be applied to obtain masks with values in $[0,1]$.

\subsubsection{Branch-Specific Edge Masks}

For a branch $b=(b_1,b_2,\dots,b_N)$, define the set of high-frequency axes as
\begin{equation}
    H(b)=\{i\in\{1,2,\dots,N\}: b_i=H\}.
\end{equation}

For example,
\begin{equation}
    H(LHH)=\{2,3\},
    \qquad
    H(LLL)=\emptyset.
\end{equation}

For branches with $H(b)\neq\emptyset$, define the branch-specific edge mask as
\begin{equation}
    M_b(\mathbf{x})
    =
    \operatorname{Norm}
    \left(
    \sqrt{
    \sum_{i\in H(b)}
    \left\|
    \frac{\partial s(\mathbf{x})}{\partial x_i}
    \right\|_2^2
    }
    \right),
\end{equation}
where $\operatorname{Norm}(\cdot)$ denotes a normalisation operation that maps the edge magnitude to the interval $[0,1]$.

This definition assigns each high-frequency branch an edge mask corresponding to the directions in which that branch applies high-frequency coordinate scaling. The all-low-frequency branch, for which $H(b)=\emptyset$, is excluded from the edge-guidance penalty.

\subsubsection{Directional Edge-Guidance Loss}

The directional edge-guidance loss penalises high-frequency branch contributions in non-edge regions. For the $N$-dimensional case, it is defined as
\begin{equation}
    \mathcal{L}_{\mathrm{edge}}
    =
    \sum_{\substack{b\in\mathcal{B}_{N}\\H(b)\neq\emptyset}}
    \mathbb{E}_{\mathbf{x}\in\Omega_d}
    \left[
    \left(1-M_b(\mathbf{x})\right)
    \left\|
    P_b(\mathbf{x})\mathbf{y}_b(\mathbf{x})
    \right\|_1
    \right].
\end{equation}

The term $w_b(\mathbf{x})\mathbf{y}_b(\mathbf{x})$ is the actual weighted contribution of branch $b$ to the final reconstruction. Therefore, the loss penalizes high-frequency contributions only where the corresponding directional edge evidence is weak.

\subsubsection{Confidence Regularization}

In addition to penalizing the high-frequency contribution itself, we may also discourage the model from assigning high confidence to high-frequency branches in non-edge regions. This gives the confidence regularization term
\begin{equation}
    \mathcal{L}_{\mathrm{conf}}
    =
    \sum_{\substack{b\in\mathcal{B}_{N}\\H(b)\neq\emptyset}}
    \mathbb{E}_{\mathbf{x}\in\Omega_d}
    \left[
    \left(1-M_b(\mathbf{x})\right)
    P_b(\mathbf{x})
    \right].
\end{equation}

The full directional edge-guidance loss is then
\begin{equation}
    \mathcal{L}_{\mathrm{DEGL}}
    =
    \mathcal{L}_{\mathrm{edge}}
    +
    \beta_{\mathrm{conf}}
    \mathcal{L}_{\mathrm{conf}},
\end{equation}
where $\beta_{\mathrm{conf}}\geq 0$ controls the strength of the confidence regularization term.




\subsubsection{Overall Objective}

The full $N$-dimensional training objective is
\begin{equation}
    \mathcal{L}_{\mathrm{total}}
    =
    \mathcal{L}_{\mathrm{recon}}
    +
    \lambda_{\mathrm{edge}}
    \mathcal{L}_{\mathrm{DEGL}},
\end{equation}
where $\lambda_{\mathrm{edge}}\geq 0$ are the weighting coefficients for the directional edge-guidance loss and the gradient-pair loss, respectively.

\subsection{Recovery of the Two-Dimensional Case}

When $N=2$, the general branch set becomes
\begin{equation}
    \mathcal{B}_{2}
    =
    \{L,H\}^{2}
    =
    \{LL,HL,LH,HH\}.
\end{equation}

The corresponding scaling vectors are
\begin{align}
    \boldsymbol{\alpha}_{LL} &= (\alpha_L,\alpha_L),\\
    \boldsymbol{\alpha}_{HL} &= (\alpha_H,\alpha_L),\\
    \boldsymbol{\alpha}_{LH} &= (\alpha_L,\alpha_H),\\
    \boldsymbol{\alpha}_{HH} &= (\alpha_H,\alpha_H).
\end{align}

These are precisely the four branch types used in the two-dimensional image formulation: one low-frequency approximation branch, two axis-oriented high-frequency branches, and one joint high-frequency branch. The confidence-weighted reconstruction reduces to
\begin{equation}
    \hat{s}_{\theta}(\mathbf{x})
    =
    P_{LL}(\mathbf{x})\mathbf{y}_{LL}(\mathbf{x})
    +
    P_{HL}(\mathbf{x})\mathbf{y}_{HL}(\mathbf{x})
    +
    P_{LH}(\mathbf{x})\mathbf{y}_{LH}(\mathbf{x})
    +
    P_{HH}(\mathbf{x})\mathbf{y}_{HH}(\mathbf{x}).
\end{equation}

Thus, the proposed $N$-dimensional formulation is a direct extension of the implemented two-dimensional model.

\section{Remark on Computational Scalability}

The construction above is mathematically direct, but the number of branches grows exponentially with the signal dimension, since $|\mathcal{B}_N|=2^N$. Therefore, while the formulation is valid for arbitrary $N$, practical implementations for high-dimensional signals may require branch sharing, sparse branch selection, grouped frequency patterns, or other parameter-efficient approximations.

\subsection{Performance and Efficiency Ablation Study}

\begin{table}[t]
\centering
\small
\setlength{\tabcolsep}{4pt}
\begin{tabular}{lrrrrrr}
\hline
Method & P(K) & GFLOPs & Train(s) & Infer(s) & Thpt & PSNR \\
\hline
SIREN     & 199  & 25.9 & 0.222 & 0.074 & 350   & 32.9 \\
FINER     & 199  & 25.9 & 0.270 & 0.090 & 288   & 36.4 \\
INCODE    & 437  & 38.7 & 0.435 & 0.145 & 267   & 36.2 \\
WIRE      & 100  & 13.0 & 0.645 & 0.215 & 60    & 32.5 \\
COSMO RC  & 437  & 38.7 & 3.500 & 1.100 & 33.2  & 45.1 \\
FR INR    & 6294 & 29.2 & 0.229 & 0.082 & 356.1 & 36.2 \\
\hline
ScaLe INR & 534  & 69.5 & 0.820 & 0.291 & 238.8 & 46.8 \\
\hline
\end{tabular}
\vspace{2mm}
\caption{Comparison of super-resolution performance against existing SOTA methods.}
\end{table}

\paragraph{Explanation of metrics.}
\textbf{P(K)} is the number of parameters (in thousands), indicating model size. \textbf{GFLOPs} measures computational cost per forward pass. \textbf{Train(s/it)} and \textbf{Infer(s/it)} denote training and inference time per iteration, respectively. \textbf{Thpt} measures throughput in GFLOPs/s, reflecting efficiency. \textbf{+PSNR} evaluates reconstruction quality, where higher values indicate better fidelity.

\paragraph{Discussion.}
The results show a clear trade-off between efficiency and reconstruction quality. Lightweight models are computationally efficient but achieve lower PSNR, while heavier methods improve quality at higher cost. ScaLe INR achieves the best PSNR overall, outperforming all baselines while maintaining a reasonable efficiency balance compared to high-cost alternatives such as COSMO RC.

\section{Image Inpainting}
\label{res:inpainting}

\textbf{Data.}
We use the \textit{Celtic Spiral Knots} image with resolution $572 \times 582 \times 3$. A random mask is applied such that only $20\%$ of pixel coordinates are observed during training. Evaluation is performed by reconstructing the full image grid.

\textbf{Implementation Details.}
Training uses 500 epochs with batch size $256 \times 256$, learning rate $1.5 \times 10^{-4}$, and decay factor $0.25$. The model requires approximately 700~MB GPU memory. A frequency scaling factor of $0.5$ is used as described in Section~\ref{method}.

\textbf{Results.}
ScaLe-INR achieves a reconstruction quality of $22.07$~dB PSNR, outperforming COSMO-RC ($21.88$~dB), INCODE ($21.85$~dB), SIREN ($20.69$~dB), FINER ($21.83$~dB), and ReLU+PE ($21.68$~dB). This consistent improvement demonstrates stronger pixel-level recovery and better structural continuity under extreme sparsity.

\begin{figure}[h]
\centering
\includegraphics[width=0.8\textwidth]{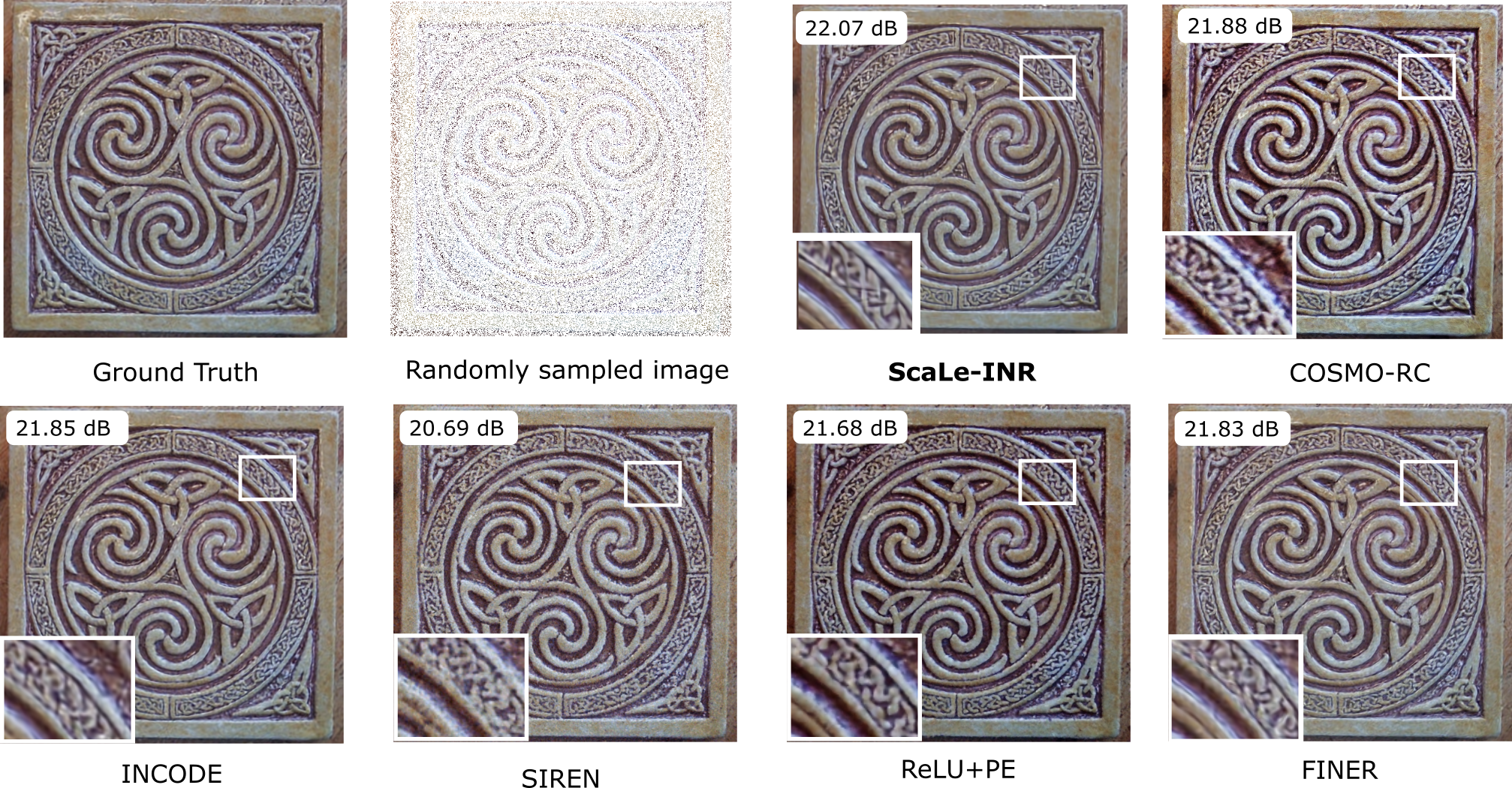}
\caption{Qualitative image inpainting results produced by ScaLe-INR.}
\label{fig:inpainting}
\end{figure}